\documentclass[sigconf]{acmart}
\renewcommand\footnotetextcopyrightpermission[1]{} % removes footnote with conference information in first column

\usepackage{fancyhdr}

\begin{document}
\pagestyle{empty}
\title{Prior Knowledge Guided Network for Video Anomaly Detection}

\author{Zhewen Deng}
\authornote{Equal contribution.}
\email{2100727@stu.edu.neu.cn}

\affiliation{%
  \institution{Northeastern University}
  \city{Shenyang}
  \country{China}
}

\author{Dongyue Chen}
\authornotemark[1]
\authornote{Corresponding author.}
\affiliation{%
  \institution{Northeastern University}
  \city{Shenyang}
  \country{China}
}

\author{Shizhuo Deng}
\affiliation{%
  \institution{Northeastern University}
  \city{Shenyang}
  \country{China}
}

\renewcommand{\shortauthors}{Deng et al.}

\begin{abstract}
Video Anomaly Detection (VAD) involves detecting anomalous events in videos, presenting a significant and intricate task within intelligent video surveillance. Existing studies often concentrate solely on features acquired from limited normal data, disregarding the latent prior knowledge present in extensive natural image datasets. To address this constraint, we propose a Prior Knowledge Guided Network(PKG-Net) for the VAD task. First, an auto-encoder network is incorporated into a teacher-student architecture to learn two designated proxy tasks: future frame prediction and teacher network imitation, which can provide better generalization ability on unknown samples. Second, knowledge distillation on proper feature blocks is also proposed to increase the multi-scale detection ability of the model. In addition, prediction error and teacher-student feature inconsistency are combined to evaluate anomaly scores of inference samples more comprehensively. Experimental results on three public benchmarks validate the effectiveness and accuracy of our method, which surpasses recent state-of-the-arts.
\end{abstract}

\maketitle

\section{Introduction}
Due to the wide use of monitoring systems, detecting anomalous events (e.g., crime, violent fight, car accident) from surveillance videos has become a significant task. The considerable quantity of surveillance cameras in the real world makes it impossible for human to process all the video data in real-time. Hence, automatic anomaly detection is an essential and challenging topic in computer vision and pattern recognition. On the one hand, anomalous events rarely happen in the real world compared with normal events. On the other hand, anomalous events cannot be defined previously, making it unlikely to collect all kinds of anomalous events. Therefore, the training set only contains normal samples, while abnormal samples only appear in the testing set in most anomaly detection datasets. Consequently, anomaly detection is considered an unsupervised or self-supervised task, and the typical solution~\cite{chen2022comprehensive,liu2021hybrid,park2020learning} is to propose proxy tasks that perform well on normal samples while badly on abnormal ones.

Existing works for anomaly detection can be classified into two categories: generation-based method and feature-based method. The generation-based method usually uses an encoder-decoder network to generate reconstruction~\cite{nguyen2019anomaly} or prediction~\cite{liu2018future} of given frames. The model only trained on normal data is supposed to perform badly on abnormal data, which leads to larger reconstruction errors. Recent studies following this approach~\cite{yu2020cloze,liu2021hybrid} attempt to increase the difficulty of the proxy task to prevent methods from possessing excessive ability to reconstruct or predict abnormal frames. To achieve this goal, two-stream architecture~\cite{yu2020cloze}, memory-augmented structure~\cite{park2020learning}, image masking strategy~\cite{zhong2022cascade}, etc. are introduced to limit the generalization ability of the generation-based models. However, without considering semantic, high-dimensional criteria, these methods may still fail in certain cases inevitably.

Typical feature-based methods~\cite{kwon2020backpropagated,yi2020patch} project input images into a high-dimensional feature space where normal and abnormal samples can be distinguished easily according to a probability distribution, distance measurements, or supervised classification. Recent works~\cite{bergmann2020uninformed,wang2021glancing} train the feature network to learn the proxy task of teacher imitation on industrial images~\cite{bergmann2019mvtec}, in which knowledge is transferred from pre-trained teacher networks~\cite{simonyan2014very,he2016deep} to relatively shallow student networks through knowledge distillation~\cite{hinton2015distilling}. Since the model is trained only on normal samples, a larger feature discrepancy between the student and the teacher networks can be expected for abnormal samples. These models work better in detecting large-scale abnormal targets while ignoring finer-scale anomalies because the knowledge distillation is commonly conducted on the output layers of the feature networks.

\begin{figure*}[t]
\centering
\includegraphics[width=2.0\columnwidth]{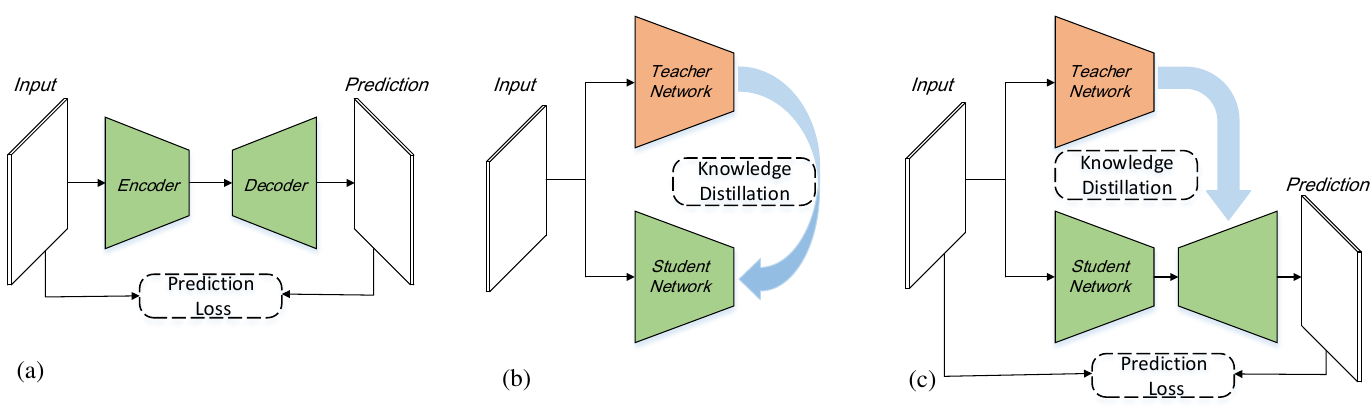}
\caption{(a) FFP predicts future frames using an encoder-decoder network. (b) KD transfers knowledge from a teacher network to a student network. (c) Our PKG-net predicts future frames under the guidance of prior knowledge learned from the teacher network.\label{figure1}}
\end{figure*}

We propose a simple but effective architecture, referred to as Prior Knowledge Guided Network (PKG-Net), to integrate generation-based and feature-based frameworks. First, an encoder-decoder network is embedded into a teacher-student architecture to learn two proxy tasks: future frame prediction and teacher network imitation, which is expected to learn low-level features from normal data while extracting the high-level knowledge representation from the expert teacher network. Specifically, we conduct knowledge distillation on proper blocks of the teacher network to select suitable feature scales for the VAD task. For inference samples, prediction error and teacher-student feature inconsistency are combined to evaluate the anomaly scores. In summary, our contributions can be concluded as follows:

\begin{itemize}
\item We propose a new baseline for video anomaly detection referred to as PKG-Net by embedding a future frame prediction network into a teacher-student architecture, which can detect more varieties of anomalies with more stable generalization ability.
\item We present a block-wise knowledge distillation strategy for the proposed PKG-Net to achieve anomaly detection on proper scales.
\item Experimental results demonstrate the effectiveness and accuracy of our method, which surpass the previous state-of-the-art.
\end{itemize}

\section{Related Work}
\textbf{Future frame prediction(FFP).} An unsupervised VAD prototype adopts generative models for frame generation. FFP \cite{liu2018future} is a typical and widely used VAD model along this approach, whose architecture is roughly shown in Fig. \ref{figure1}(a). FFP utilizes an Unet \cite{ronneberger2015u} structure to predict future frames and recognize frames with larger prediction errors as anomalies. FFP hypothesizes that the model trained on normal data can only successfully predict normal frames but fail on abnormal frames. However, this assumption is sometimes invalid when the model is trained perfectly to have great generalization ability on anomaly samples as well. Conversely, an overfitted model will give false alarms frequently because it may recognize normal samples that have not been seen before as anomalies. In a word, it is difficult for a typical generation-based VAD model to balance its generalization ability. Besides, the encoder-decoder structure pays more attention to low-level features, which limit the ability of the model to detect anomalies at different scales.

\noindent\textbf{Knowledge distillation(KD).} Knowledge distillation for anomaly detection can be categorized into the feature-based approach, which was first introduced by Uniformed-students \cite{bergmann2020uninformed} to detect anomalies from industrial images. Different from classic feature-based methods, KD recognizes anomalies according to the teacher-student feature inconsistency. As shown in Fig. \ref{figure1} (b), a teacher-student structure transfers knowledge from an expert teacher network to a student network by applying knowledge distillation on normal data, while the larger difference in outputs between the student and the teacher networks can be expected for anomaly samples because the shallow structure of the student makes it difficult to mimic the teacher on anomaly samples. Along this approach, a global-local comparison model \cite{wang2021glancing} is further developed to train a local network and a global network through knowledge distillation and then detect anomalies according to the global-local feature inconsistency. Although KD-based methods inherit semantic knowledge and high-level features from the pre-trained model and consequently have better generalization ability, they are still limited by lack of low-level visual features, improper selection of feature scales, and the bias between the training data of the application scenario and the large-scale image datasets for teacher network pre-training.

To deal with the problem mentioned above, we design the PKG-Net as shown in Fig. \ref{figure1} (c) to acquire high-level knowledge from the teacher network and predict the frames at pixel level simultaneously, which can provide a stronger ability of anomaly detection on proper scales and a better balance of generalization ability between normal and abnormal data.

\section{Methodology}
In the following sections, we present the principle of our proposed PKG-Net first, then discuss the detailed structure of the PKG-Net and finally present the loss function at the training stage and the way to evaluate anomaly score at the inference stage.

\begin{figure*}[ht]
\centering
\includegraphics[width=2.03\columnwidth]{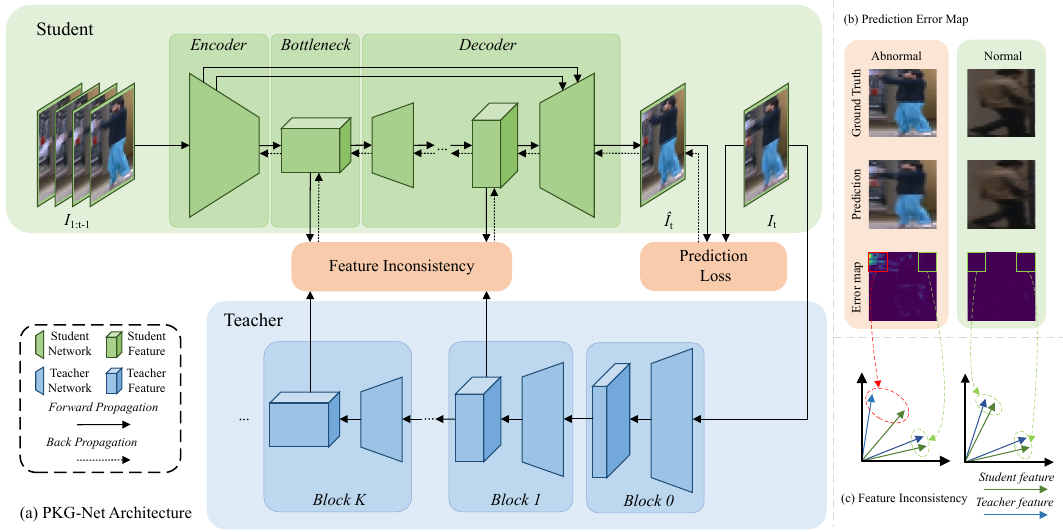}
\caption{The pipeline of our proposed PKG-net. (a) The student auto-encoder network takes the previous frames $I_{1:t-1}$ as the inputs and gives a prediction $\hat{I}_{t}$ for the future frame $I_t$. The pre-trained teacher network takes the original future frame $I_{t}$ as the input to produce high-level features. During the training, knowledge is transferred from certain blocks of the teacher network to the student network by minimizing feature inconsistency loss. While the pixel-level MSE loss is used to reduce the prediction error of $\hat{I}_{t}$. (b) The pixel-level prediction error of anomalous samples is higher than that of normal samples under the guidance of prior knowledge. (c) The teacher-student feature inconsistency of anomalous samples is higher than that of normal samples. \label{figure2}}
\end{figure*}

\subsection{Architecture and motivations}
FFP is supposed to predict the future frames of normal samples with lower error while predicting anomalous samples with larger error, in which the pixel-level Mean Square Error(MSE) loss is used to minimize the prediction error. Since more attention is paid to details of the predicted frames, the large-scale and high-level anomalies are likely to miss. Besides, the generalization ability of such generation models is difficult to control. The anomalies can be missed because the model with a strong generalization ability will produce a small prediction error even for anomalous samples. In contrast, normal samples can be misidentified as anomalies if the model is overfitted. To overcome the above limitations, we propose a new framework for VAD, referred to as Prior Knowledge Guided Network(PKG-Net), which imports high-level features and semantic representation from a pre-trained teacher network using knowledge distillation. The architecture of the PKG-Net is illustrated in Fig.\ref{figure2}(a).

In the PKG-Net, the student is aimed to predict normal frames and imitate the teacher to acquire the representing ability on normal data. During inference, samples with higher feature inconsistency or larger prediction errors are more likely to be anomalies. Compared with the traditional criterion of prediction error, feature inconsistency between student and teacher networks offers a high-level criterion for large-scale anomalies. The combination of the two criteria, of which the nature is two different proxy tasks: predicting future frame and imitating teacher, enables the model to detect multi-scale anomalies. In addition, the generalization ability of the model can be better balanced by learning the two proxy tasks at the same time. On the one hand, the generalization on unknown normal data can be further improved under the guidance of prior knowledge because the student inherits the semantic representation of normal data from the teacher. On the other hand, the integration of the two proxy tasks makes it difficult for anomalous samples to get small prediction errors and low feature inconsistency simultaneously, in the sense that the generalization on anomalies is correspondingly limited. Consequently, the generalization ability of the proposed framework can be controlled properly.
\vspace{-0.3cm}

\subsection{Detailed Structure}
The primary issue of the proposed PKG-Net is the selection of the teacher network and the construction of the student network. The teacher network should produce high-level features and semantic representations for nature images in order to guide the student network. Therefore, ResNet \cite{he2016deep} and its variants \cite{xie2017aggregated,zagoruyko2016wide} are good candidates for the teacher network as they are capable of extracting rich features from nature images(see bottom row of Fig.\ref{figure2}). The student network is supposed to predict future frames while possessing the structure for learning high-level representation from the teacher network. The student structure, as shown in the top row of Fig.\ref{figure2}, adopts an auto-encoder with skip connections to achieve frame prediction and feature learning. Another challenge of the proposed method is the relationship between student and teacher for knowledge distillation. Knowledge distillation is conducted on specific blocks of the teacher network whose output feature scale is suitable for the current VAD task. As the encoder corresponds to the previous frames while the bottleneck and the decoder are related to the predicted frame, knowledge distillation is carried out from certain blocks of the teacher to the bottleneck and decoder of the student. Notably, the feature tensor of the bottleneck part and the decoder in the student network is exactly the same size as the feature tensor of certain blocks in the teacher network.

Unet-like skip connections are adopted in the student network to preserve low-level features for the proxy task of frame prediction. As the low-level features are mostly acquired by learning the prediction task on normal data, detailed anomalies can be detected easily. However, if the low-level feature is learned from the teacher network, anomalies in finer scales may not be recognized due to the model may predict abnormal frames properly with the feature of good generalization ability. Regarding the proxy task of knowledge distillation, high-level semantic features are transferred from the teacher to the student, which covers the shortage in feature scales of the prediction model. Moreover, the student network generalizes better on unseen normal data with the prior knowledge inherited from the teacher network. The combination of the two proxy tasks helps the student network to detect anomalies in multiple scales. In addition, the generalization ability is also constrained appropriately, which improves the robustness and universality of VAD. Please refer to our appendix for more details on the network design.

\subsection{Loss Function}
As mentioned above, the loss function of our proposed PKG-Net is composed of two parts: future frame prediction error and teacher-student feature inconsistency. Correspondingly, the anomaly score of an input sample can be computed based on the two factors.

\noindent\textbf{Prediction Loss.}
To make the prediction close to its ground truth, we minimize the pixel-level MSE loss between the predicted frame $\hat{I}_{t}$ and its ground truth $I_{t}$:
\begin{equation}
    L_{\rm{e}}=\Vert I_{t}-\hat{I}_{t}\Vert^{2}_{2}.
\end{equation}
In addition, a widely used loss term, referred to as Gradient loss \cite{mathieu2015deep}, is also introduced to sharp the predicted frame, as shown as follows:
\begin{equation}
\begin{aligned}
 L_{\rm{g}}(\hat{I_{t}},I_{t})&=\sum_{i,j}\left|\lvert I_{i,j}-I_{i-1,j}\rvert -\lvert \hat{I}_{i,j}-\hat{I}_{i-1,j}\rvert\right|^{\alpha}\\
    &+\left|\lvert I_{i,j-1}-I_{i,j}\rvert-\lvert\hat{I}_{i,j-1}-\hat{I}_{i,j}\rvert\right|^{\alpha},
\end{aligned} 
\end{equation}
where $i,j$ denote the spatial pixel index of a frame, and $\alpha\geq1$ is an integer adjusting the sharpness degree of the prediction.

\noindent\textbf{Feature Inconsistency.}
The cosine similarity, which captures the relation of feature representation, is applied along the channel axis to measure the feature inconsistency between the two feature vectors $\boldsymbol{f}_{\rm{s}}^{k}$ and $\boldsymbol{f}_{\rm{t}}^{k}$ at the position $(m,n)$ of $k$-th blocks of the student and the teacher networks, as shown as follows:
\begin{equation}
    {l}_{\rm{c}}(m,n,k) = 1-\frac{(\boldsymbol{f}_{\rm{s}}^{k})^{\rm{T}}\boldsymbol{f}_{\rm{t}}^{k}}{\Vert \boldsymbol{f}_{s}^{k}\Vert\Vert \boldsymbol{f}_{t}^{k}\Vert}.
\end{equation}
The total feature inconsistency loss is calculated by averaging the vector-wise cosine similarity over the $K$ blocks:
\begin{equation}
    L_{\rm{c}} = \frac{1}{K} \sum_{k=1}^{K}\left[\frac{1}{M_{k}N_{k}}\sum_{m=1}^{M_{k}}\sum_{n=1}^{N_{k}}{l}_{\rm{c}}(m,n,k)\right], 
\end{equation}
where $M_{k},N_{k}$ indicate the height and width of the feature tensor of feature vectors of the $k$-th block, respectively. Together, the overall loss function of the PKG-Net takes the form as follows:
\begin{equation}
    L=\lambda_{\rm{e}}L_{\rm{e}}+\lambda_{\rm{g}}L_{\rm{g}}+\lambda_{\rm{c}}L_{\rm{c}}, 
\end{equation}
where $\lambda_{\rm{e}}, \lambda_{\rm{g}}$ and $\lambda_{\rm{c}}$ are the corresponding weights to balance different loss terms, respectively.

\subsection{Anomaly Scoring}
The anomaly score takes a similar form as the loss function and consists of two parts: prediction error and block-wise feature inconsistency.

The prediction error $S_{\rm{e}}$ between the predicted frame and the ground truth can be calculated as follows: 
\begin{equation}
    S_{\rm{e}}=\Vert \hat{I}_{t}-I_{t}\Vert^{2}_{2}.
\end{equation}

The inconsistency $S_{\rm{c}}^{k}$ of the $k$-th block between the student network and teacher network takes the form of the following equation:
\begin{equation}
    S_{\rm{c}}^{k}=\frac{1}{M_{k}N_{k}}\sum_{m=1}^{M_{k}}\sum_{n=1}^{N_{k}}\left[1-\frac{(\boldsymbol{f}_{\rm{s}}^{k})^{T}\boldsymbol{f}_{\rm{t}}^{k}}{\Vert \boldsymbol{f}_{\rm{s}}^{k}\Vert\Vert \boldsymbol{f}_{\rm{t}}^{k}\Vert}\right].
\end{equation}

We combine these two parts to achieve the final score as follows:
\begin{equation}
    S=w_{\rm{e}}\frac{S_{\rm{e}}-\mu_{\rm{e}}}{\sigma_{\rm{e}}}+\sum_{k=1}^{K}\left(w_{\rm{c}}^{k}\frac{S_{\rm{e}}^{k}-\mu_{\rm{c}}^{k}}{\sigma_{\rm{c}}^{k}}\right),
\end{equation}
where $w_{\rm{e}}$ and $w^{k}_{\rm{c}}$ are the weights for prediction error and feature inconsistency. $\mu_{\rm{e}}$, $\mu^{k}_{\rm{c}}$, $\sigma_{\rm{e}}$ and $\sigma^{k}_{\rm{c}}$ are the means and standard deviations of prediction error and the feature inconsistency of the $k$-th block on all training samples respectively, which can be computed from the training set and applied in the testing. It is easy to understand that the larger the value of $S$, the higher the risk of anomaly.

\section{Experiments}
\subsection{Experimental Settings}

\noindent\textbf{Evaluation Metrics.} Following the previous works \cite{liu2021hybrid,chen2022comprehensive}, we measure the Area Under the Receiver Operation Characteristic (AUROC) by varying the threshold over the anomaly score for evaluation. Higher AUROC indicates better performance of anomaly detection.

\noindent\textbf{Datasets.} Following the widely used prototypes, the proposed model is evaluated on three benchmark datasets for video anomaly detection, including UCSD Ped2 \cite{mahadevan2010anomaly}, CUHK Avenue \cite{lu2013abnormal} and ShanghaiTech \cite{luo2017revisit}. \textbf{UCSD ped2} dataset contains 16 training and 12 testing video clips with pedestrian movement parallel to the camera plane. The dataset has 2550 training frames and 2010 testings, with a spatial resolution of 320x240. Testing video samples contain anomalous events such as skateboarding, cycling, vehicles appearing on the sidewalk, etc. \textbf{CUHK Avenue} dataset is composed of 16 training videos with 15328 frames and 21 testing ones with 15324 frames. These frames are acquired by a fixed 150 fps camera, and each frame size is 360x640. The dataset includes 47 anomalies such as running, throwing bags, walking in the wrong direction, etc. \textbf{ShanghaiTech} is the largest dataset for video anomaly detection so far, with up to 274k training frames and 42k testing frames from 13 different scenarios. It contains 130 abnormal events, including violent fights, chasing, robbing, etc.

\begin{figure}[t]
\centering
\includegraphics[width=1.0\columnwidth]{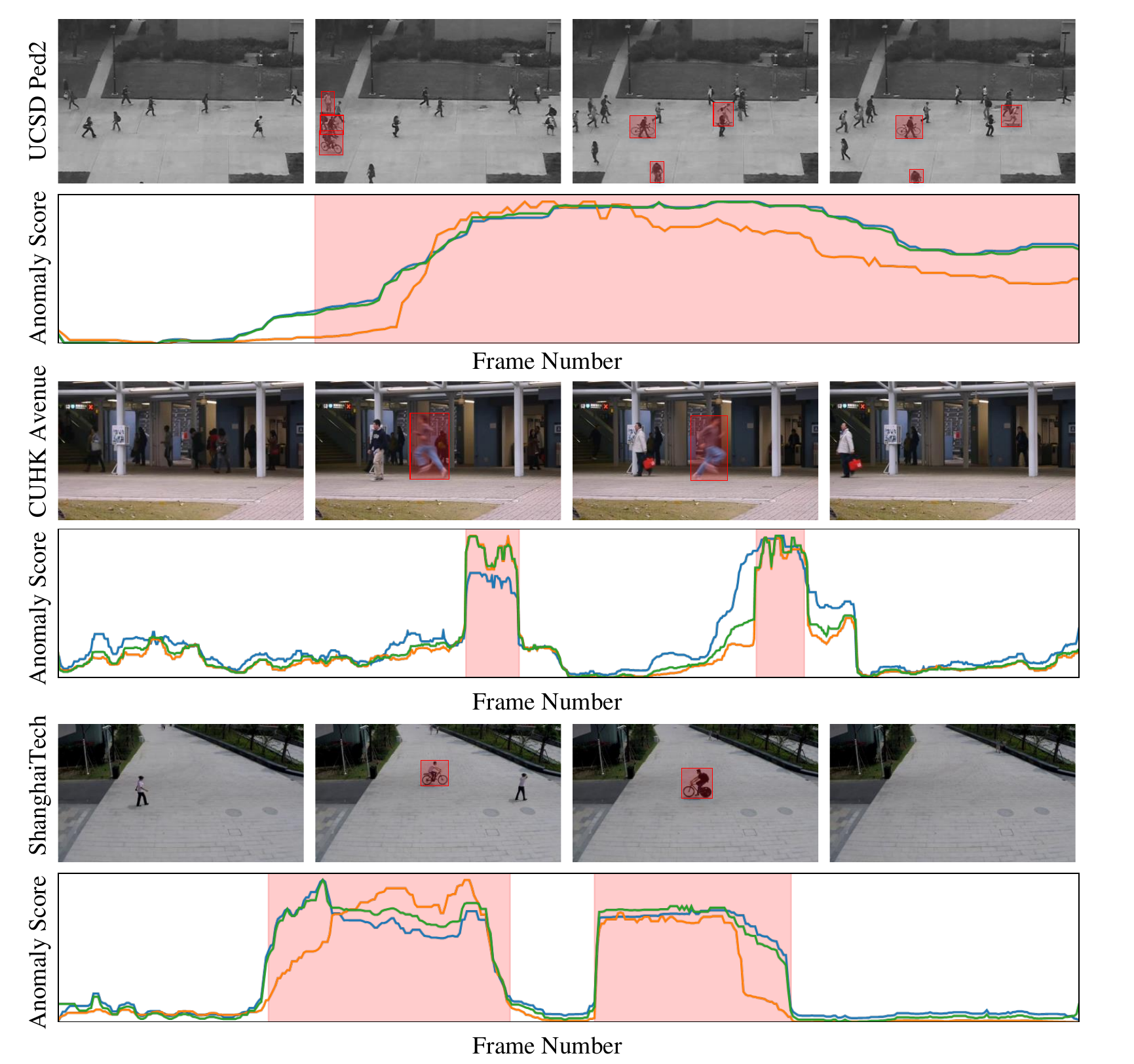}
\caption{Anomaly score curves of testing videos on three benchmarks. Blue, orange, and green curves stand for prediction error, feature inconsistency, and the combined anomaly score, respectively. Abnormal intervals are marked in red.\label{figure5}}
\end{figure}

\noindent\textbf{Implementation Details.} Following \cite{yu2020cloze,chen2022comprehensive}, we train our model on the foreground objects instead of the whole frames or frame patches, which can be extracted by pre-trained Cascade RCNN \cite{cai2018cascade} and then resized to $32\times32$. A spatial-temporal cube (STC) containing each foreground object of the current frame and the previous $i$ frames is made as the input of the model (typically $i$=4). The maximum anomaly score of all objects in one frame is selected to be the anomaly score of the whole frame for CUHK Avenue and ShanghaiTech. For UCSD ped2, we calculate the average of the top-3 anomaly scores in one frame as the frame anomaly score because multiple anomaly objects frequently appear in the same frame in the dataset (see Fig. \ref{figure5}). Following \cite{liu2021hybrid}, we smooth the anomaly scores of a video with a median filter considering the continuity of the events.

We adopt ResNext50 as the teacher network for CUHK Avenue and ResNet50 for UCSD Ped2 and ShanghaiTech, respectively. The proposed framework is implemented with the PyTorch library and the Adam optimizer with $\beta=(0.9,0.999)$. The initial learning rate is  $1e^{-4}$, decayed by 0.8 after every 60 epochs. $\lambda_{\rm{e}}$, $\lambda_{\rm{g}}$, $\lambda_{\rm{c}}$ are $0.7$, $0.1$, $0.2$, respectively. The batch size and epoch number of UCSD Ped2, CUHK Avenue, and ShanghaiTech are set to (128,120), (128,120), and (256,80). Specifically, we set $K$=2 for UCSD Ped2 and $K$=1 for CUHK Avenue and ShanghaiTech. The weights of anomaly score for UCSD Ped2, CUHK Avenue and ShanghaiTech are set to ($w_{\rm{c}}^{1}$=0.65, $w_{\rm{c}}^{2}$=0.35, $w_{\rm{p}}$=0.01), ($w_{\rm{c}}^{1}$=0.5, $w_{\rm{p}}$=0.5), and ($w_{\rm{c}}^{1}$=0.15, $w_{\rm{p}}$=0.85), respectively.

\subsection{Ablation Study}

\begin{table}[b]
 \small%\footnotesize
     \caption{Ablation on knowledge distillation of single blocks.}
 \setlength{\belowcaptionskip}{0.5cm}
  \centering

  \setlength{\tabcolsep}{5mm}{
  %\scalebox{2.0}{
    \begin{tabular}{l|l|l|l}
        \hline
        Block & Ped2 & Avenue & SHTech \\
        \hline
        Block1 & 91.0\% & 75.0\% & 76.5\% \\
        Block2 & \textbf{98.1}\% & \textbf{83.4}\% & \textbf{77.6}\% \\
        Block3 & 97.9\% & 81.4\% & 76.6\% \\
        Block4 & 84.4\% & 78.1\% & 72.7\% \\
        \hline
    \end{tabular}}%

  \label{table2}%
\end{table}%

\textbf{Block settings for knowledge distillation.}
To analyze the influence of block settings on the overall performance of the proposed model, we carried out knowledge distillation on different blocks of the Resnet50. One group of experiments is implemented by only applying knowledge distillation on different single blocks. In the default setting, the bottleneck of the student network corresponds to Block2. For the experiments on Block3 and Block4, specifically, the bottleneck of the student is set as Block3 and Block4, respectively. From the results in Tab.\ref{table2}, we can see that the knowledge distillation on Block2 achieves the best performance. Block1 is not suitable due to the weaker representation ability of its lower-level features. The typical KD for industrial image anomaly detection(i.e. Block4 KD) has a significant AUC degradation because its preferred scales are too coarse for the targets in surveillance videos. Though block3 achieves decent results as well, the limited size of the bottleneck slightly damages the prediction ability of the student network and consequently decreases the overall performance.

\begin{table}[h]
 \small%\footnotesize
     \caption{Ablation on knowledge distillation of combined blocks.}
 \setlength{\belowcaptionskip}{0.5cm}
  \centering

  \setlength{\tabcolsep}{4mm}{
  %\scalebox{2.0}{
    \begin{tabular}{l|l|l|l}
        \hline
        KD setting& Ped2 & Avenue & SHTech \\
        \hline
        K=1(Block2) & 98.7\% & \textbf{93.8\%} & \textbf{80.2\%} \\
        K=2(Block12) & \textbf{99.5\%} & 93.1\% & 78.2\% \\
        K=3(Block123) & 99.0\% & 90.6\% & 76.6\% \\
        \hline
    \end{tabular}}

  \label{table3}%
    \vspace{-0.3cm}
\end{table}%

The other group of experiments is conducted on the three benchmarks to find the best combination of blocks for knowledge distillation in PKGNet. As shown in Tab.\ref{table3}, The scheme with Block1 and Block2 distilled achieves the best on UCSD Ped2 while the one only with Block2 achieves better results on Avenue and ShanghaiTech. One obvious reason is that the targets in Ped2 are uniformly smaller and concentrate on a finer scale, while those of Avenue and ShanghaiTech are commonly bigger and have a wider distribution on different scales. Though the addition of Block3 also achieves reasonable results, the narrow bottleneck of the student network still limits the prediction of future frames. Thus, the best block settings are Block1+Block2 for Ped2 and only Block2 for CUHK Avenue and ShanghaiTech, respectively. All these results reveal that different block settings for knowledge distillation have a remarkable impact on the overall performance due to their various preference of feature scales, and our work can provide a compatible framework for various block settings of knowledge distillation.

\vspace{-0.4cm}
\begin{table}[h]
 \small%\footnotesize
     \caption{Ablation on different backbones of teacher model.}
 \setlength{\belowcaptionskip}{0.5cm}
  \centering

  \setlength{\tabcolsep}{4mm}{
  %\scalebox{2.0}{
    \begin{tabular}{l|l|l|l}
    \hline
    Teacher model& Ped2 & Avenue & SHTech \\
    \hline
    ResNet18 & 98.8\% & 83.4\% & 77.3\% \\
    ResNet50 & \textbf{99.5\%} & \textbf{93.8\%} & 78.3\% \\
    ResNext50 & 99.0\% & 92.4\% & \textbf{80.2\%} \\
    WResNet50 & 99.1\% & 92.0\% & 78.0\% \\
    \hline
    \end{tabular}}

  \label{table4}%
    \vspace{-0.3cm}
\end{table}%

\textbf{Backbone of teacher network.} The knowledge contained in the teacher network can be varied with different backbones, which may lead to a change in performance. Quantitative comparisons between ResNet, ResNext, and wide-ResNet as the teacher network are given in Tab.\ref{table4}. ResNet50 and ResNext50 achieve the best results on different datasets, respectively. Notably, the lightweight backbone with ResNet18 as the teacher also achieves the competitive result on UCSD Ped2 and ShanghaiTech.

\begin{table}[h]
 \small%\footnotesize
     \caption{Ablation on architectures.}
 \setlength{\belowcaptionskip}{0.5cm}
  \centering

  \setlength{\tabcolsep}{4mm}{
  %\scalebox{2.0}{
    \begin{tabular}{l|l|l|l}
        \hline
        Architecture & Ped2 & Avenue & SHTech \\
    
        \hline
        AE & 92.9\% & 89.7\% & 74.1\% \\
        KD & 98.1\% & 83.4\% & 77.6\% \\
        PKG-Net & \textbf{99.5}\% & \textbf{93.8\%} & \textbf{80.2\%} \\
        \hline
    \end{tabular}}

  \label{table5}%
    \vspace{-0.3cm}
\end{table}%

\textbf{Architecture of PKG-Net.}
To verify the effectiveness of the proposed PKG-Net, we compared our method with a naive auto-encoder(AE) and the knowledge distillation(KD) method on the three datasets. The experimental results are shown in Tab.\ref{table5}. For Ped2, ShanghaiTech, the anomalies are much determined by the appearance features at the semantic level; therefore, KD is a better alternative than AE. For Avenue, most anomaly samples stand for unusual actions which are unfamiliar even to the teacher model; therefore, AE beats KD on this task. In addition, the proposed PKG-Net outperforms AE by evident improvements of 6.6\%, 4.1\%, 6.1\% on Ped2, Avenue, and ShanghaiTech, respectively. It also achieves significant gains of 1.4\%, 10.4\%, 2.6\%  in comparison with using KD alone. All the above results indicate that the proposed PKG-Net can integrate the advantages of both KD and AE and provides a better anomaly detection solution than using KD or AE separately.

\subsection{Comparison with the State-of-the-art}

\noindent\textbf{Quantitative Results.}
We compared our PKG-Net with the state-of-the-art methods, including reconstruction-based, prediction-based, hybrid and other categories. As shown in Tab.\ref{table1}, our method creates a new paradigm for prediction-based approaches and achieves the SOTA performance on three datasets. Since both prediction error and teacher-student feature inconsistency are considered in the proposed architecture to control the generalization ability properly, and the suitable feature scales are selected by implementing knowledge distillation with appropriate block settings, the performance of anomaly detection is improved on the three benchmarks, respectively, compared with the previous best SOTA methods. Notably, the proposed method is optical-flow free, while most of the previous SOTA methods need to compute optical-flow features, which makes our method more competitive in both computational cost and accuracy. Moreover, our method can achieve competitive results if used in a reconstruction way. Please refer to our appendix for such results.

\begin{table}[b]

 \small%\footnotesize
     \caption{Comparison between our method and recent state-of-the-art methods.}
  \centering

  \setlength{\tabcolsep}{3.0mm}{
  %\scalebox{2.0}{
    \begin{tabular}{llll}
    \hline
    \textbf{Method} & \textbf{Ped2} & \textbf{Avenue} & \textbf{SHTech} \\
    \hline
    \textbf{Reconstraction-based}\\
    ConvLSTM-AE\cite{luo2017remembering} & 88.1\% & 77.0\% & - \\
    MemAE\cite{gong2019memorizing} & 94.1\% & 83.3\% & 71.2\% \\
    \hline
    \textbf{Prediction-based}\\
    FFP\cite{liu2018future} & 95.4\% & 85.1\% & 72.8\% \\
    VEC\cite{yu2020cloze} & 97.3\% & 90.2\% & 74.8\% \\
    Bi-direction\cite{chen2022comprehensive} & 98.3\% & 90.3\% & 78.1\% \\
    \hline
    \textbf{Hybird and others}\\
    AMMC-Net\cite{cai2021appearance} & 96.6\% & 86.6\% & 73.7\% \\
    HF2VAD\cite{liu2021hybrid} & \underline{99.3\%} & 91.1\% & 76.2\% \\
    USTN-DSC\cite{yang2023video} & 98.1\% & 89.9\% & 73.8\% \\
    HSNBM\cite{bao2022hierarchical} & 95.2\% & 91.6\% & 76.5\% \\
    DERN\cite{sun2022evidential} & 97.1\% & \underline{92.7\%} & \underline{79.3\%} \\
    \hline
    \textbf{PKG-Net(Ours)} & \textbf{99.5}\% & \textbf{93.8\%} & \textbf{80.2\%}\\
    \hline
    \end{tabular}}

  \label{table1}%
    \vspace{-0.4cm}
\end{table}%

\begin{figure}[t]
\centering
\includegraphics[width=\columnwidth]{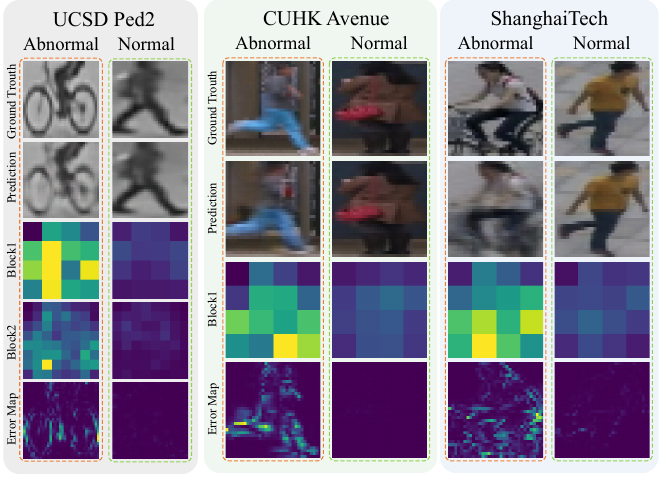}
\caption{Visualization of feature inconsistency and prediction error maps for our method. From top to bottom, we show ground-truth frames, predicted frames, feature inconsistency maps of distilled blocks, and prediction error maps. The brighter colors denote larger values of feature inconsistency and prediction error.}\label{figure6}
\end{figure}

\noindent\textbf{Visualization.} To further illustrate the mechanism of our method, several maps of feature inconsistency and prediction error for some typical samples in UCSD Ped2, CUHK Avenue, and ShanghaiTech are presented in Fig. \ref{figure6} respectively. It can be seen that the feature inconsistency and prediction errors increase dramatically for abnormal samples, which means both the two proxy tasks, teacher imitation, and future frame prediction, are effective in distinguishing anomalies from normal samples. Another important point worth mentioning is that our method demonstrates different preferences for anomalies in different scales, which can be observed from the various patterns of the feature inconsistency maps and the prediction error maps. Specifically, feature inconsistency maps are more likely to highlight the anomalies in larger scales, such as bicycle wheels, body movements, and the interactions between targets and environments, while prediction error maps pay more attention to boundaries, moving edges, or finer-scale textures. Thus, the combination of the two discrepancies produces a stronger multi-scale detection ability, enabling our model to recognize more variety of visual anomalies.

\section{Conclusions}
This paper proposes a novel framework PKG-Net with an integrated architecture of an auto-encoder and a pre-trained CNN network to realize anomaly detection based on two proxy tasks: future frame prediction and teacher imitation. Specifically, the AE network is trained to predict future frames based on its previous frame sequence while imitating the pre-trained teacher network with knowledge distillation on proper feature blocks. In addition to prediction error, teacher-student feature inconsistency is also leveraged as a new criterion for video anomaly, improving the multi-scale detection ability and the generalization ability of the model. Extensive experiments illustrate that our method achieves superior results over the previous state-of-the-art methods.


\end{document}